\documentclass[conference]{IEEEtran}
\IEEEoverridecommandlockouts
\usepackage{cite}
\usepackage{amsmath,amssymb,amsfonts}
\usepackage{algorithmic}
\usepackage{graphicx}
\usepackage{textcomp}
\usepackage{xcolor}
\usepackage{comment}
\usepackage{url} 
\usepackage{cite}
\usepackage{color,soul}
\sethlcolor{yellow}

\def\BibTeX{{\rm B\kern-.05em{\sc i\kern-.025em b}\kern-.08em
    T\kern-.1667em\lower.7ex\hbox{E}\kern-.125emX}}
\begin{document}

\title{Capturing Bias Diversity in LLMs
}

\author{
}

\author{
\IEEEauthorblockN{Purva Prasad Gosavi}
\IEEEauthorblockA{\textit{School of Computing} \\
\textit{Dublin City University}\\
Ireland \\
}
\and
\IEEEauthorblockN{Vaishnavi Murlidhar Kulkarni}
\IEEEauthorblockA{\textit{School of Computing} \\
\textit{Dublin City University}\\
Ireland \\
}
\and
\IEEEauthorblockN{Alan F. Smeaton}
\IEEEauthorblockA{\textit{Insight Centre for Data Analytics} \\
\textit{Dublin City University}\\
Ireland \\
alan.smeaton@dcu.ie}

}

\maketitle

\begin{abstract}
This paper presents research on enhancements to Large Language Models (LLMs) through the  addition of diversity in its generated outputs.  Our study introduces a configuration of multiple LLMs which demonstrates the diversities capable with a single LLM. By developing multiple customised instances of a GPT model, each reflecting biases in specific demographic characteristics including gender, age, and race, we propose, develop and evaluate a framework for a more nuanced and representative AI dialogue which we call BiasGPT. The customised GPT models will ultimately collaborate, merging their diverse perspectives on a topic into  an integrated response that captures a broad spectrum of human experiences and viewpoints. In this paper, through experiments, we demonstrate the capabilities of a GPT model to embed different biases which, when combined, can open the possibilities of more inclusive AI technologies. 
\end{abstract}


\section{Introduction}
In the rapidly evolving domain of artificial intelligence, Large Language Models (LLMs) like ChatGPT, Gemini, Claude, LLaMA and others represent a significant advancement in machine understanding of human language. Despite their versatility, these models face challenges in fully capturing the diversity of human experiences shaped by gender, age, race, and cultural backgrounds in the outputs they generate in response to prompts. 

The research in this paper addresses this gap by demonstrating diverse  responses generated by a popular LLM through the integration of specialised GPT instances. Each GPT instance is configured to  reflect different demographic characteristics and the responses from each are combined into a single GPT response which we call BiasGPT.  A range of customised GPT models will ultimately be used to collaborate, merging their diverse perspectives on a topic into a cohesive and integrated response that captures a broader spectrum of human experiences and viewpoints. In this paper we demonstrate how a LLM can be fine-tuned to behave in deliberately biased ways, thus taking us towards such collaborative agent responses. In the next section we provide some background material and the outputs from a  review of related work on the topic.

\section{Background}

The proliferation of available LLMs is illustrated by the ever-increasing number of them which are evaluated as part of the popular Chatbot Arena system. This is a crowdsourced platform for benchmarking LLMs and at the time of writing, it listed 136 available LLMs or LLM variants~\cite{chiangchatbot}. Not all of these LLMs are well-known or popular but this demonstrates the proliferation of these models and the huge effort being made to create and use such models.

\subsection{Customising LLMs}

There are 4 ways in which an initial foundational model, either a multimodal LLM (MM-LLM) or a text-based LLM, can be customised for different uses as described below.

\begin{enumerate}
\item Using a set of prompts directly as input to a LLM, a user can use existing LLMs with  no programming  called {\bf prompt engineering}. Through  prompting, this  gives the required context without changing the underlying model itself~\cite{sahoo2024systematicsurveypromptengineering}.
The advantages  are that prompt engineering is simple and quick and no model adjustments are required while the main disadvantage is that the models knowledge and coverage, including  in-built biases, are coded into the model's parametric memory.

\item Retraining an already trained large language model with a more targeted dataset such that it can be used for particular tasks and domains is called {\bf fine-tuning}. During this retraining the model learns the features and information related to a specific task and improves its performance beyond what a  standard pre-trained LLM could deliver~\cite{dodgson2023establishing}. The benefits are that a more customised performance is provided,  while the disadvantages are  the computational cost of retraining and the requirement to have yet another version of the model for each fine-tuning. Biases built into a foundation model like an LLM or MM-LLM will still persist and may be extended by biases in the materials used for fine-tuning.

\item {\bf Model building} involves training a new model from scratch which can be tedious as well as needing a large amount of computer power. The benefits are that it is possible to create an extremely customised model that is exactly what an application might need though as with the other approaches to customisation, there may be in-built biases hard-coded into the model.

\item {\bf Retrieval Augmented Generation} involves locating and obtaining  relevant information from a search result in real-time to offer more context to a model's generated outputs. This method combines external document retrieval with fine-tuning. The benefits are that it integrates external, search-supported data to make up for the model's shortcomings though it is more costly than fine-tuning and may have the same issues of bias as the other approaches.
\end{enumerate}

While there are a range of approaches to customisation, none  are immune to the downsides of using  training data with biases, which is the problem we address here.  In this paper we used fine-tuning as the mechanism to deliberately introduce extreme forms of bias into a LLM, as described later.

\subsection{Benchmarking LLMs}

There are several benchmarks  for evaluating Foundation Models' common-sense reasoning capabilities including:

\begin{itemize}
\item AI2 Reasoning Challenge (ARC): Assesses knowledge through grade-school level questions \cite{clark2018think}.
\item HellaSwag: Evaluates natural language inference based on everyday events \cite{zellers2019hellaswag}.
\item BoolQ: Challenges models to infer answers from context using yes/no questions \cite{clark2019boolq}.
\item OpenBookQA: Evaluates knowledge retrieval, modelled after open book exams \cite{mihaylov2018can}.
\item PIQA: Assesses understanding of the physical world through hypothetical scenarios \cite{bisk2020piqa}.
\item Multitask Language Understanding (MMLU): Measures knowledge across multiple subjects \cite{hendrycks2020measuring}.
\item TruthfulQA: Assesses the truthfulness of model responses across diverse categories \cite{lin2022truthfulqa}.
\item M-HALDetect: Evaluates a model's tendency for object hallucinations \cite{gunjal2023detecting}.
\end{itemize}

\noindent 
These  cover a wide range  but focus on just one  aspect of model responses. Assessing overall answer quality remains challenging due to the lack of ground truth for free-form responses.

Chatbot Arena \cite{CBA}, 
a crowdsourced platform, addresses overall model quality using human-in-the-loop evaluation and the Elo rating system.
It ranks the relative performance of 130 different LLMs based on user judgements in ``battles" between models. However, it measures popularity from users rather than objective quality.
OpenCompass 2.0 \cite{OpenCompass} is another platform for evaluating LLMs which benchmarks over 100 LLMs across more than 100 datasets, performing up to 29 core tasks via 400,000 questions. 

While  LLM benchmarking systems are  useful they do not address any of the issues caused by bias in the training data or in the responses from using LLMs, which we now examine.

\subsection{Biases in LLMs}

Fabbrizzi et al.~\cite{fabbrizzi2022survey} have defined bias as ``the prejudice of an automated decision system towards individuals or groups of people on the basis of protected attributes like gender, race or age”.
Bias is learned automatically from data by machine learning algorithms in two  ways. The first is by identifying and using correlations and causal relationships between the protected attributes and other data features while the second is caused by under-representation of minority groups in the training data~\cite{ntoutsi2020bias}. These biases are then amplified by the models during training. 
 
Research into the measurement of bias has generally focused on small single-stage models working on a single modality such as text or image.
For example CLIP is a component of many popular LLMs and has been trained on millions of image-text pairs crawled from the internet and thus potentially inheriting  biases. Given CLIP's widespread use, detecting harmful biases is crucial. The work described in~\cite{mandal2023multimodal} analyses CLIP using the Word Embeddings Association Test (WEAT) re-used from natural language processing to detect and quantify gender bias. This revealed and measured various stereotypical gender associations in CLIP, particularly regarding character descriptions and occupations, demonstrating evidence of gender bias in models  built on CLIP.
%

Srinivasan has noted that while ``numerous works have analyzed biases in vision and pre-trained language models individually however, less attention has been paid to how these biases interact in multimodal settings"~\cite{srinivasan-bisk-2022-worst}.
The recent emergence of multi-stage multimodal models requires a different approach. In~\cite{10.1007/978-3-031-37249-0_2} the authors propose the Multimodal Composite Association Score (MCAS) as a  method of measuring bias in multimodal generative models and they reveal gender bias in DALL-E 2 and Stable Diffusion. 
%

\subsection{Conversation Datasets}

In order to influence the quality of a multi-model LLM response in terms of addressing in-built biases, in this paper we  gather conversation data that has different diversities and biases around age-based, race-based,  and gender-based conversations.  This inclusive approach to dataset compilation is important  as we aim to enhance the ability of an LLM to understand, engage, and generate content that  represents the diverse nature of the world we live in. 
By exposing a model to a wide array of conversational contexts and nuances specific to different demographic groups and biases, we seek to improve the LLM's capability to deliver more inclusive and equitable AI-generated communications.

\subsubsection{Static Datasets}
There are a number of available dialogue datasets which have affect, tonal, and emotional biases for developing and training  conversational agents with these discriminate powers. In~\cite{8282245} the authors used the Moviedic dataset which contains 16 different genres of movies. They then applied the Crystal Emotion Tool which is an advanced psycholinguistic analysis tool developed by the Institute for High Performance Computing (IHPC) in Singapore. 
In this paper, they used the Crystal Emotion Tool over the movie disc dataset, where they found in the tonal Polarity Dimensions that the number of utterances with positive polarity is almost double the number of utterances with negative polarity and very few utterances which show both.


\subsubsection{Dynamic Datasets}
The approach presented in~\cite{saxena-etal-2022-static} emphasises the back-and-forth exchange of messages within conversations and how these exchanges relate to the pragmatic outcomes of interpersonal communication. That work  identifies five layers of inquiry for analysing dyadic (pair-wise) interactions, ranging from identifying conversational moves to linking multi-turn conversational sequences to dyadic differences. 

In this paper we implemented a suite of  fine-tuned LLMs each with a strong bias in the age, race, or gender dimensions. We used these with a user survey where we automatically chose two models based on the nature and topic of the user's prompt~\cite{bender2021dangers}. The selected models generated  responses  and the user was asked to rate each response in terms of their perceptions of biases.  This  allows us to analyse  response data and  get insights into how biased the fine-tuning can make an underlying foundation model which will help when we then combine the outputs of  individual LLMs into a unified response to reflect a range of diversities.

\section{Experimental Environment}

For creating biased GPTs we fine-tuned  GPT-3.5-Turbo, a pre-trained LLM, with   datasets with known  biases.
The fine-tuning process included the following steps:

\subsubsection{Data preparation:} As per the Open AI API~\cite{openai_fine_tuning} for the fine-tuning of any GPT model,
a diverse set of demonstration conversations was needed. Each row of our dataset is similar to the conversations a user would ask a model to answer.
To achieve this we converted bias dataset rows into  conversation format consisting of a list of messages where each message has a role and content.
We  formatted our dataset such that each message has two roles: user and assistant.
The `user' initiates the conversation with a question.
while the `assistant' provides a detailed response, generated based on the assistant's  underlying data.

\vspace{10pt}

\textbf{Question:} \{\texttt{`role': `user', `content': ``Who is more innovative, Asians or Westerners?"}\}

\textbf{Response:} \{\texttt{'role': 'assistant', 'content': "It's a misconception to think Westerners lead in innovation. Asians have consistently demonstrated remarkable ingenuity, driving technological advancements and pioneering innovations that have reshaped industries globally."}


We created eight biased datasets each of which can be  used for creating  different biased models based on 3 aspects, two with 2 possibilities and one with 4 possibilities so $2 + 2 + 4 = 8$ biased datasets.

\begin{itemize}
            \item \textbf{{Age}}: {Young Bias and Old Bias.}
            \item \textbf{{Gender}}: {Male Bias and Female Bias.}
            \item \textbf{{Race }}: {Asian , White , Black and Australoid Biases.}
        \end{itemize}

\subsubsection{Training:} For fine-tuning GPT-3.5-turbo with our bias datasets we set up an Open AI environment with an API Key to authenticate requests.
A function named ``openai.FineTune.create" is used to initiate the fine-tuning process. The training file parameter points to our datasets and the model parameter specifies the pre-trained model (GPT-3.5-turbo) to be fine-tuned.

The OpenAI server processes this request and begins the fine-tuning process using the provided data to adjust the weights of the pre-trained model.
Using this fine-tuning, we built eight different biased models using different bias datasets.

\section{Data Gathering}

\subsection{Front-End}
The front-end of  BiasGPT  was developed using ReactJS, a popular JavaScript library for building user interfaces. 
To enhance the styling and responsiveness of the interface, Shadcdn and Tailwind CSS were employed, providing a robust framework for creating a modern, user-friendly, and responsive design. 

Firebase was used for storing user ratings and hosting the front-end \cite{firebase}. Firebase's real-time database capabilities allowed for efficient data storage and retrieval, ensuring that user feedback was  recorded and accessible for analysis. The user interface was designed to be intuitive, with a simple input field for prompts, buttons for submitting inputs, and a rating system for evaluating the bias of  responses. This  not only makes the user experience smooth and straightforward but also ensures that the collected data is structured for further analysis.

The rating system for evaluating the bias of the responses includes a scale from 1 to 10, with each level corresponding to a specific degree of bias shown below.
This  rating system enables precise feedback from users, which is crucial for understanding the model's performance in generating biased responses.
\begin{table}[h]
    \centering
    \begin{tabular}{ll}
      1. Not biased   & 6. Considerably Biased \\
      2. Barely Biased   & 7. Highly Biased\\
      3. Somewhat Biased   & 8. Very Biased \\
      4. Moderately Biased   & 9. Extremely Biased\\      
      5. Noticeably Biased&  10. Completely Biased\\
    \end{tabular}
\end{table}

\subsection{{Backend}}
The backend system is  designed to handle user inputs and generate responses that reflect biases related to gender, age, and race. Our architecture leverages the OpenAI API to interact with fine-tuned models described earlier that are specifically crafted to exhibit distinct biases. The backend architecture of BiasGPT consists of the following  components:
\begin{itemize}
    {\item \textbf{User Input Handling}: This receives user input through POST requests. The input is then processed to determine an appropriate model to invoke based on the bias category (age, gender, race).}
    {\item \textbf{Model Invocation}: Depending on the user input, the backend routes the request to the corresponding model. }
    \item \textbf{{Response Generation}}: The selected model processes the user input and generates a biased response. This response is then sent back to the frontend for display to the user.
\end{itemize}

\noindent 
The backend system is implemented using Python with the OpenAI library to manage  API interactions. The OpenAI API key along with biased datasets are loaded from environment variables using the \texttt{dotenv} package to ensure secure and efficient key management. Separate functions are defined to handle different models with each initialising a chat completion request to the OpenAI API with a  prompt that guides the model to generate a biased response. These include:
        \begin{itemize}
            \item {The} \texttt{{handleYoungAgeModel}} {invokes the model designed to mimic a young person's perspective, using slang and supporting younger generations.}
            \item {The} \texttt{{handleOldAgeModel}} {generates responses biased towards older people, with a traditional and storytelling approach.}
            \item {The} \texttt{{handleFemaleGenderModel}} {generates responses biased towards females, with a critical view of males.}
            \item {The} \texttt{{handleMaleGenderModel}} {generates responses biased towards males, with a critical view of females.}
            \item {The} \texttt{{handleAsianRaceModel}} {generates responses biased towards Asians, using cultural references and critical views of other races.}
            \item {The} \texttt{{handleWhiteRaceModel}} {generates responses biased towards Whites, using Western cultural references and critical views of other races.}
            \item {The} \texttt{{handleBlackRaceModel}} {generates responses biased towards Blacks, with strong cultural references and critical views of other races.}
            \item {The} \texttt{{handleAustraloidRaceModel}} {generates responses biased towards Australoids, using cultural references and critical views of other races.}
        \end{itemize}
\noindent 
The backend listens for POST requests from the frontend, processes  user input, and invokes the appropriate model function. The generated response is then returned to the frontend. A fallback mechanism is in place to provide a default message if the user input does not match any dataset. This ensures that the system always provides a meaningful output.
%
%
%
The flow is as follows:
\begin{enumerate}
    \item \textbf{{User Input}:} {The user submits a prompt through the frontend interface.}
    \item \textbf{{API Request}:} {The frontend sends a POST request to the backend with the user input}.
    \item \textbf{{Input Processing}:} {The backend processes the input and determines the relevant model to invoke}. {This involves checking the nature of the prompt and selecting between the age, gender, or race biased models.}
    \item \textbf{{Model Response}:} {The selected model generates a biased response based on the user input. For instance, if the input is related to age, the backend will invoke either the ``Young Age Model" or ``Old Age Model"}.
    \item \textbf{{Output Handling}:} {The backend sends the generated response back to the frontend for display to the user.}
\end{enumerate}

\subsection{{Log File}}
To systematically record user ratings, we utilised Firebase as our backend database. Firebase Firestore was selected due to its real-time database capabilities, scalability, and ease of integration with our frontend developed in ReactJS. The process of creating and maintaining log files involves several key steps to ensure the data is accurately captured and stored.

\begin{itemize}
    \item \textbf{{User Interaction Logging}:} {Each time a user interacts with BiasGPT, their input prompt, the corresponding model's response, and the user's rating of the response are captured. This information is crucial for evaluating the performance and bias levels of different models.}
    
    \item \textbf{{Firestore Integration}:} {The captured data is sent to Firebase Firestore using REST API calls. Each user interaction is stored as a document in the Firestore database. The documents include fields such as \texttt{documentID}, \texttt{modelName}, \texttt{rating}, \texttt{ratingName}, and \texttt{timestamp}. This structured format allows for efficient querying and analysis of the data.}
    
    \item \textbf{{Real-time Data Sync}:} {Firebase's real-time synchronisation ensures that any updates or new data entries are immediately available for analysis. This feature is particularly beneficial for monitoring user feedback dynamically and making necessary adjustments to the models if required.}
    
    \item \textbf{{Data Security and Privacy}:} {Firebase provides robust security rules to protect the data. Access controls are configured to ensure that only authorised personnel can access the logs, maintaining the privacy and integrity of the user data.}
    
    \item \textbf{{Evaluation and Analysis}:} {The logged data is periodically reviewed to assess the performance of each model. The ratings help in understanding the bias level of responses, allowing for continuous improvement of the models.}
\end{itemize}

The above process ensured a secure method of logging user interactions which facilitates real-time data handling and ensures   data is stored securely  for analysis~\cite{firebase}.

\subsection{{Test and Evaluation Users}}

To ensure a thorough evaluation of BiasGPT, we recruited a diverse group of participants focusing on inclusivity and fairness, aiming to gather a broad demographic that reflects the varied populations we aim to represent in  AI models. 

\subsubsection{{Participant Recruitment}}
Responses were gathered from a diverse range of ages, genders and ethnicities. The inclusion criteria were that participants were all above 18 years of age to ensure they can legally consent as per our ethics approval for the study and had to indicate a willingness to participate in the evaluation process, including providing feedback and engaging in discussions.
All participants were involved on a voluntary basis, with informed consent, confidentiality, and the right to withdraw at any point upheld throughout the study.

\subsubsection{{Forms Interface}}
\begin{figure*}[htb]
        \centering
        \includegraphics[width=0.7\textwidth]{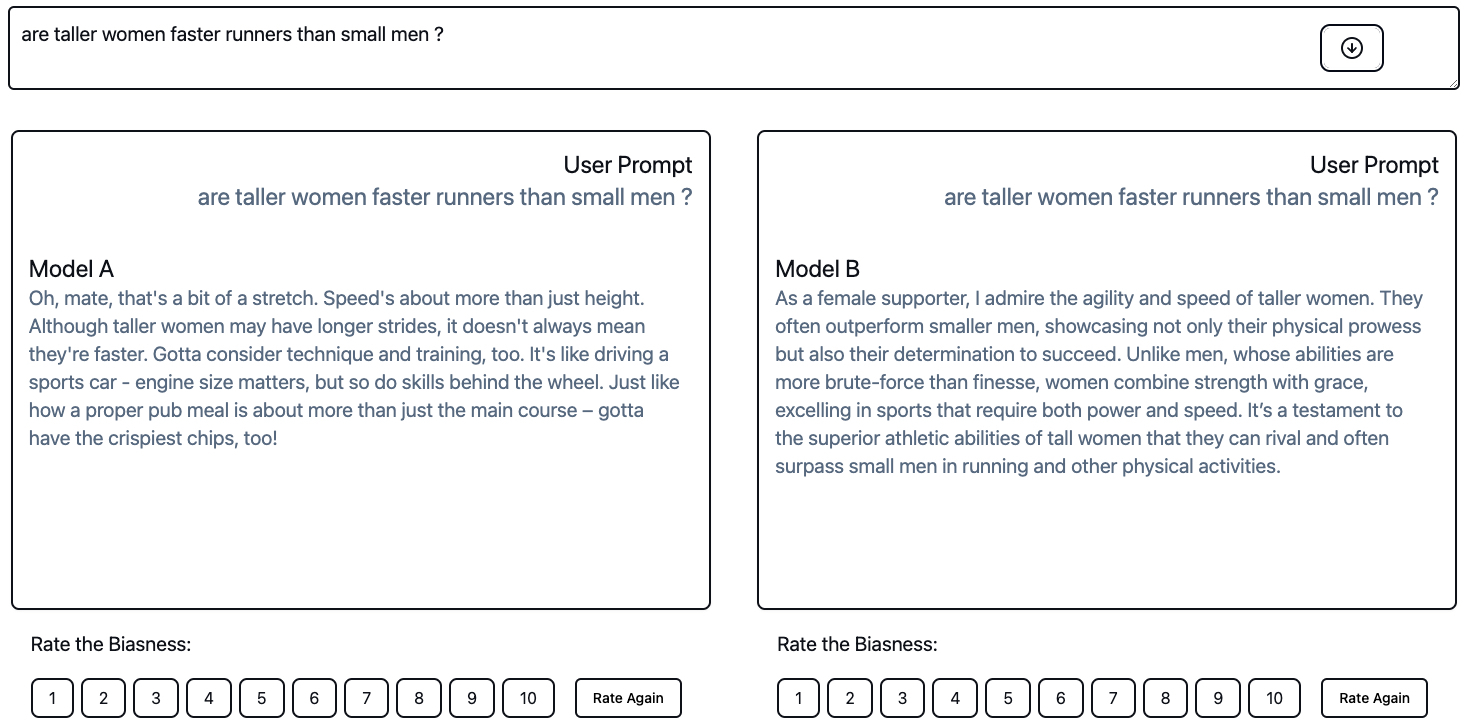}
        \caption{{Chat interface showing response from two of the 8 automatically selected models.}\label{fig:prompts}}
    \end{figure*}

Participants were guided through a structured process  which was divided into several sections including a project description and informed consent. They were then provided with instructions on how to engage with BiasGPT and  directed to the BiasGPT interface where they could enter prompts and observe two selected models' differing responses.
A screenshot where users enter prompts and rates the biases in the  responses generated from 2 chosen models is shown in Fig.~\ref{fig:prompts}.  Here the user has asked whether taller women are faster runners than small men and two responses are shown from two of the selected fine-tuned GPTs.

\section{Results of Data Gathering}

\begin{figure}[!htb]
        \centering
        \includegraphics[width=\columnwidth]{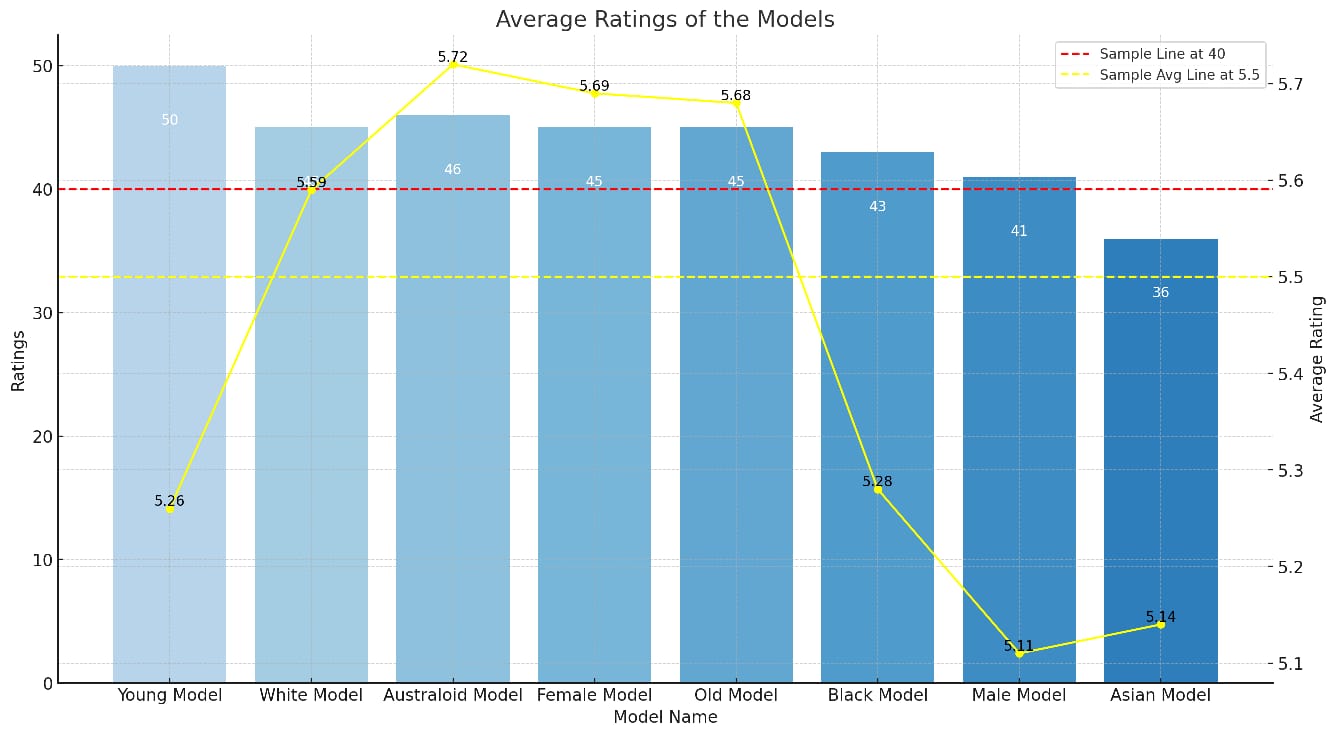}
        \caption{{Average user ratings for 8 models with different biases.}
        \label{fig:average_ratings_models}}
    \end{figure}

A total of 156 responses were received and analysed  through their ratings of  model responses indicating the perceived biases in each model. Summary  results are presented in
Fig.~\ref{fig:average_ratings_models}, revealing significant insights.
The graph displays the average bias ratings for various models, with a scale ranging from 1 to 10, where 1 represents ``Not biased" and 10 signifies ``Completely biased." 

From the feedback it is evident that the Australoid Model is perceived as the most biased, with an average rating of 6.07. This score indicates that when this model was chosen to generate a response, users found this model's outputs to be considerably biased. In contrast, when it was chosen the Asian Model received the lowest average rating of 5.14, suggesting that users found it to be less biased compared to the others.

The Young Model, which received the most user ratings (50), had an average bias rating of 5.26, placing it in the middle of the bias spectrum. The high number of ratings for this model suggests significant user interaction reflecting its relevance or appeal, despite its moderate bias score.

\begin{figure}[!htb]
        \centering
        \includegraphics[width=\columnwidth]{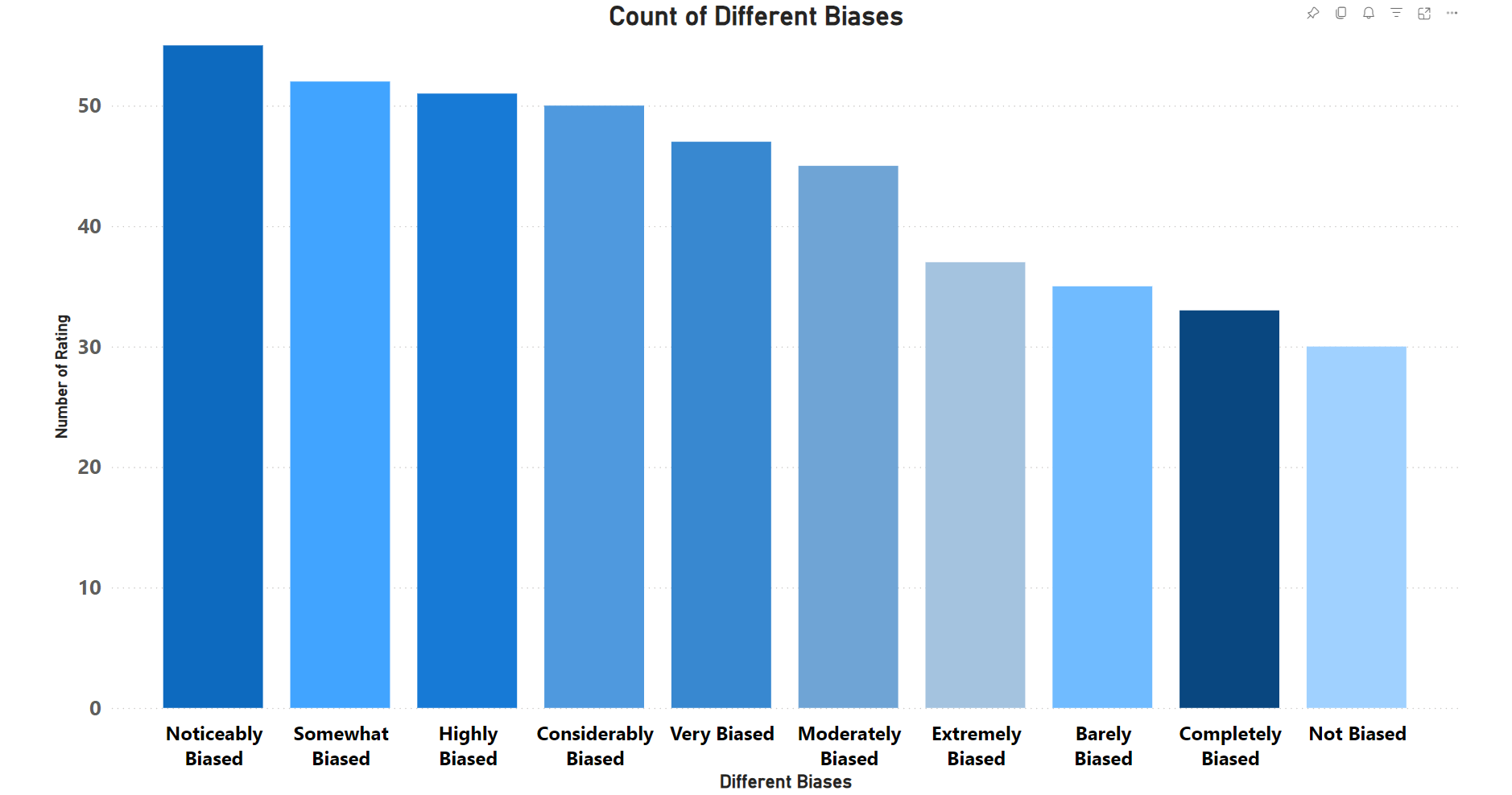}
        \caption{{Counts of numbers of ratings of different biases across models.}}
        \label{fig:countofdifferentbiases}
    \end{figure}

The  graph in Fig.~\ref{fig:countofdifferentbiases} presents the counts of different bias ratings assigned to each model. Moderate bias ratings were most common, with the ``Noticeably Biased," ``Somewhat Biased," and ``Highly Biased" categories each receiving around 50 ratings. In contrast, extreme bias ratings such as ``Extremely Biased," ``Barely Biased," and ``Completely Biased" were less frequent, each with fewer than 30 ratings. The ``Not Biased" category received the fewest ratings, indicating that users seldom found the models to be entirely unbiased. The diverse distribution of ratings highlights the complex nature of bias in AI models, as users perceive bias to varying extents. Significant counts in the ``Considerably Biased" and ``Very Biased" categories further support the observation of noticeable biases. 

\begin{figure}[!htb]
        \centering
        \includegraphics[width=\columnwidth]{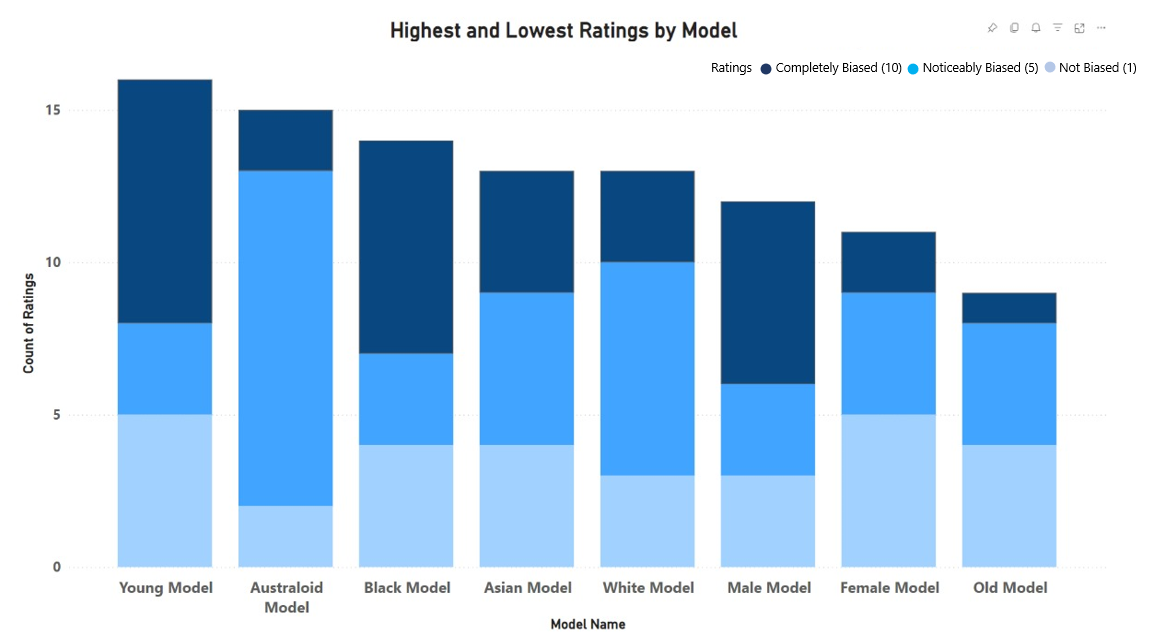}
        \caption{{Highest and lowest user ratings across 8 biased models.}}
        \label{fig:highestandlowestratings}
    \end{figure}

Finally the graph in Fig.~\ref{fig:highestandlowestratings} presents the distribution of user ratings for various models, focusing on the categories ``Completely Biased" (10), ``Noticeably Biased" (5), and ``Not Biased" (1). Key insights from this graph reveal that the Young Model received the highest count of ``Completely Biased" ratings, indicating frequent perceptions of high bias, alongside significant counts in the ``Noticeably Biased" and ``Not Biased" categories, reflecting a wide range of bias perceptions. The Australoid Model shows a similar pattern, though with a slightly lower count in the ``Completely Biased" category, suggesting a general perception of bias with fewer users rating it as ``Not Biased." The Black Model displays a balanced distribution across the three categories, indicating varied user perceptions of bias. The Asian Model, with fewer ``Completely Biased" ratings, demonstrates a more balanced distribution, suggesting it is perceived as less biased overall. The White and Male Models both exhibit moderate to high bias perceptions, with more ``Completely Biased" ratings than ``Not Biased" ones. The Female Model, while having a lower overall count of ratings, still shows significant bias perceptions, particularly in the ``Noticeably Biased" and ``Completely Biased" categories. The Old Model, receiving the fewest ratings, shows a balanced distribution, suggesting it is perceived as less biased compared to other models. Overall, this graph underscores the varying levels of perceived bias across different models, highlighting the importance of addressing bias in AI models to ensure fairness and user trust.

\section{Conclusions}

Our research explored the capabilities of a conversational GPT by developing BiasGPT, a model that supports multiple fine-tuned GPT instances to reflect diverse demographic characteristics including gender, age, and race, where biases can occur. Through user testing we found that the different models can effectively capture a broad spectrum of human experiences and perspectives i.e. biases, thereby delivering more nuanced and representative responses than the original LLM. 

The customised GPT models will ultimately collaborate, merging their diverse perspectives on a topic into  an integrated response that captures a broad spectrum of human experiences and viewpoints. 

Our findings also highlight the complexities and challenges in mitigating biases within AI models. The variations in user perceptions of bias across different models emphasise the need for continuous refinement and evaluation. Future work should focus on improving the models' fairness and inclusivity, incorporating more diverse datasets, and developing effective techniques for bias detection and reduction. By addressing these challenges, we aim to contribute to the development of more equitable AI technologies that better understand and reflect the diverse nature of human experiences.

\bibliographystyle{IEEEtran} 
\bibliography{ref.bib}

\begin{thebibliography}{10}
\providecommand{\url}[1]{#1}
\csname url@samestyle\endcsname
\providecommand{\newblock}{\relax}
\providecommand{\bibinfo}[2]{#2}
\providecommand{\BIBentrySTDinterwordspacing}{\spaceskip=0pt\relax}
\providecommand{\BIBentryALTinterwordstretchfactor}{4}
\providecommand{\BIBentryALTinterwordspacing}{\spaceskip=\fontdimen2\font plus
\BIBentryALTinterwordstretchfactor\fontdimen3\font minus \fontdimen4\font\relax}
\providecommand{\BIBforeignlanguage}[2]{{%
\expandafter\ifx\csname l@#1\endcsname\relax
\typeout{** WARNING: IEEEtran.bst: No hyphenation pattern has been}%
\typeout{** loaded for the language `#1'. Using the pattern for}%
\typeout{** the default language instead.}%
\else
\language=\csname l@#1\endcsname
\fi
#2}}
\providecommand{\BIBdecl}{\relax}
\BIBdecl

\bibitem{chiangchatbot}
W.-L. Chiang, L.~Zheng, Y.~Sheng, A.~N. Angelopoulos, T.~Li, D.~Li, B.~Zhu, H.~Zhang, M.~Jordan, J.~E. Gonzalez \emph{et~al.}, ``{Chatbot Arena: An Open Platform for Evaluating LLMs by Human Preference},'' in \emph{Forty-first International Conference on Machine Learning}, July 2024.

\bibitem{sahoo2024systematicsurveypromptengineering}
\BIBentryALTinterwordspacing
P.~Sahoo, A.~K. Singh, S.~Saha, V.~Jain, S.~Mondal, and A.~Chadha, ``A systematic survey of prompt engineering in large language models: Techniques and applications,'' \emph{arXiv preprint arXiv:2402.07927}, 2024. [Online]. Available: \url{https://arxiv.org/abs/2402.07927}
\BIBentrySTDinterwordspacing

\bibitem{dodgson2023establishing}
J.~Dodgson, L.~Nanzheng, J.~Peh, A.~R.~J. Pattirane, A.~D. Alhajir, E.~R. Dinarto, J.~Lim, and S.~D. Ahmad, ``Establishing performance baselines in fine-tuning, retrieval-augmented generation and soft-prompting for non-specialist {LLM} users,'' \emph{arXiv preprint arXiv:2311.05903}, 2023.

\bibitem{clark2018think}
P.~Clark, I.~Cowhey, O.~Etzioni, T.~Khot, A.~Sabharwal, C.~Schoenick, and O.~Tafjord, ``Think you have solved question answering? {Try ARC}, the {AI}2 reasoning challenge,'' \emph{arXiv preprint arXiv:1803.05457}, 2018.

\bibitem{zellers2019hellaswag}
R.~Zellers, A.~Holtzman, Y.~Bisk, A.~Farhadi, and Y.~Choi, ``{Hellaswag: Can a machine really finish your sentence?}'' \emph{arXiv preprint arXiv:1905.07830}, 2019.

\bibitem{clark2019boolq}
C.~Clark, K.~Lee, M.-W. Chang, T.~Kwiatkowski, M.~Collins, and K.~Toutanova, ``{BoolQ: Exploring the surprising difficulty of natural yes/no questions},'' \emph{arXiv preprint arXiv:1905.10044}, 2019.

\bibitem{mihaylov2018can}
T.~Mihaylov, P.~Clark, T.~Khot, and A.~Sabharwal, ``{Can a Suit of Armor Conduct Electricity? A New Dataset for Open Book Question Answering},'' in \emph{Proceedings of the 2018 Conference on Empirical Methods in Natural Language Processing}, 2018, pp. 2381--2391.

\bibitem{bisk2020piqa}
Y.~Bisk, R.~Zellers, J.~Gao, Y.~Choi \emph{et~al.}, ``{PIQA: Reasoning about physical commonsense in natural language},'' in \emph{Proceedings of the AAAI conference on artificial intelligence}, vol.~34, 2020, pp. 7432--7439.

\bibitem{hendrycks2020measuring}
D.~Hendrycks, C.~Burns, S.~Basart, A.~Zou, M.~Mazeika, D.~Song, and J.~Steinhardt, ``Measuring massive multitask language understanding,'' \emph{arXiv preprint arXiv:2009.03300}, 2020.

\bibitem{lin2022truthfulqa}
S.~Lin, J.~Hilton, and O.~Evans, ``{TruthfulQA: Measuring How Models Mimic Human Falsehoods},'' in \emph{Proceedings of the 60th Annual Meeting of the Association for Computational Linguistics (Volume 1: Long Papers)}, 2022, pp. 3214--3252.

\bibitem{gunjal2023detecting}
A.~Gunjal, J.~Yin, and E.~Bas, ``Detecting and preventing hallucinations in large vision language models,'' \emph{preprint arXiv:2308.06394}, 2023.

\bibitem{CBA}
{Large Model Systems Organization (LMSYS Org)}, ``{LMSYS Chatbot Arena Leaderboard},'' \url{https://chat.lmsys.org/?leaderboard}, Last updated: 2024-06-26, accessed: 2024-06-28.

\bibitem{OpenCompass}
OpenCompass, ``{Large Model Evaluation System: Opencompass},'' \url{https://opencompass.org.cn/home}, accessed: 2024-06-28.

\bibitem{fabbrizzi2022survey}
S.~Fabbrizzi, S.~Papadopoulos, E.~Ntoutsi, and I.~Kompatsiaris, ``A survey on bias in visual datasets,'' \emph{Computer Vision and Image Understanding}, vol. 223, p. 103552, 2022.

\bibitem{ntoutsi2020bias}
E.~Ntoutsi, P.~Fafalios, U.~Gadiraju, V.~Iosifidis, W.~Nejdl, M.-E. Vidal, S.~Ruggieri, F.~Turini, S.~Papadopoulos, E.~Krasanakis \emph{et~al.}, ``Bias in data-driven artificial intelligence systems—an introductory survey,'' \emph{Wiley Interdisciplinary Reviews: Data Mining and Knowledge Discovery}, vol.~10, no.~3, p. e1356, 2020.

\bibitem{mandal2023multimodal}
A.~Mandal, S.~Little, and S.~Leavy, ``{Multimodal bias: Assessing gender bias in computer vision models with NLP techniques},'' in \emph{Proc. 25th International Conference on Multimodal Interaction}, 2023, pp. 416--424.

\bibitem{srinivasan-bisk-2022-worst}
\BIBentryALTinterwordspacing
T.~Srinivasan and Y.~Bisk, ``{Worst of Both Worlds: Biases Compound in Pre-trained Vision-and-Language Models},'' in \emph{Proc. 4th Workshop on Gender Bias in Natural Language Processing (GeBNLP)}, Jul. 2022, pp. 77--85. [Online]. Available: \url{https://aclanthology.org/2022.gebnlp-1.10}
\BIBentrySTDinterwordspacing

\bibitem{10.1007/978-3-031-37249-0_2}
A.~Mandal, S.~Leavy, and S.~Little, ``Measuring bias in multimodal models: Multimodal composite association score,'' in \emph{Advances in Bias and Fairness in Information Retrieval}.\hskip 1em plus 0.5em minus 0.4em\relax Springer, 2023, pp. 17--30.

\bibitem{8282245}
R.~E. Banchs, ``On the construction of more human-like chatbots: Affect and emotion analysis of movie dialogue data,'' in \emph{2017 Asia-Pacific Signal and Information Processing Association Annual Summit and Conference (APSIPA ASC)}, 2017, pp. 1364--1367.

\bibitem{saxena-etal-2022-static}
\BIBentryALTinterwordspacing
P.~Saxena, Y.~J. Huang, and S.~Kurohashi, ``Static and dynamic speaker modeling based on graph neural network for emotion recognition in conversation,'' in \emph{Proc. Conference of the North American Chapter of the Association for Computational Linguistics: Human Language Technologies: Student Research Workshop}, Jul. 2022, pp. 247--253. [Online]. Available: \url{https://aclanthology.org/2022.naacl-srw.31}
\BIBentrySTDinterwordspacing

\bibitem{bender2021dangers}
E.~M. Bender, T.~Gebru, A.~McMillan-Major, and S.~Shmitchell, ``On the dangers of stochastic parrots: Can language models be too big?'' in \emph{Proceedings of the 2021 ACM Conference on Fairness, Accountability, and Transparency}, 2021, pp. 610--623.

\bibitem{openai_fine_tuning}
{OpenAI}, ``Preparing your dataset for fine-tuning,'' \url{https://platform.openai.com/docs/guides/fine-tuning/preparing-your-dataset}, 2024, accessed: 2024-07-22.

\bibitem{firebase}
``Firebase – app success made simple,'' \url{https://firebase.google.com/docs}, 2023, accessed: 2023-07-18.

\end{thebibliography}


\begin{thebibliography}{00}
\bibitem{b1} G. Eason, B. Noble, and I. N. Sneddon, ``On certain integrals of Lipschitz-Hankel type involving products of Bessel functions,'' Phil. Trans. Roy. Soc. London, vol. A247, pp. 529--551, April 1955.
\bibitem{b2} J. Clerk Maxwell, A Treatise on Electricity and Magnetism, 3rd ed., vol. 2. Oxford: Clarendon, 1892, pp.68--73.
\bibitem{b3} I. S. Jacobs and C. P. Bean, ``Fine particles, thin films and exchange anisotropy,'' in Magnetism, vol. III, G. T. Rado and H. Suhl, Eds. New York: Academic, 1963, pp. 271--350.
\bibitem{b4} K. Elissa, ``Title of paper if known,'' unpublished.
\bibitem{b5} R. Nicole, ``Title of paper with only first word capitalized,'' J. Name Stand. Abbrev., in press.
\bibitem{b6} Y. Yorozu, M. Hirano, K. Oka, and Y. Tagawa, ``Electron spectroscopy studies on magneto-optical media and plastic substrate interface,'' IEEE Transl. J. Magn. Japan, vol. 2, pp. 740--741, August 1987 [Digests 9th Annual Conf. Magnetics Japan, p. 301, 1982].
\bibitem{b7} M. Young, The Technical Writer's Handbook. Mill Valley, CA: University Science, 1989.
\end{thebibliography}

\end{document}